\documentclass[final]{cvpr}

\usepackage{times}
\usepackage{epsfig}
\usepackage{graphicx}
\usepackage{amsmath}
\usepackage{amssymb}

\usepackage[pagebackref=true,breaklinks=true,colorlinks,bookmarks=false]{hyperref}

\usepackage[utf8]{inputenc} %
\usepackage[T1]{fontenc}    %
\usepackage{url}            %
\usepackage{booktabs}       %
\usepackage{amsfonts}       %
\usepackage{nicefrac}       %
\usepackage{microtype}      %

\usepackage{subcaption}
\usepackage{graphicx}
\usepackage{amsmath}
\usepackage{xcolor}

\graphicspath{{images/}}

\newcommand{\picSmall}[3]{
	\begin{figure}[tb] 
		\centering
		\includegraphics[width=#3\linewidth]{#1}
		\caption{#2}
		\label{fig:#1}
	\end{figure}
}

\newcommand{\picTwo}[4]{
	\begin{figure*}[tb]
		\centering
		\begin{subfigure}[c]{0.49\textwidth}
			\centering		
			\includegraphics[width=0.9\textwidth]{#1-0}
			\subcaption{#2}	
			\label{fig:#1-0}	
		\end{subfigure}
		\begin{subfigure}[c]{0.49\textwidth}	
			\centering	
			\includegraphics[width=0.9\textwidth]{#1-1}
			\subcaption{#3}
			\label{fig:#1-1}	
		\end{subfigure}
		\caption{#4}
		\label{fig:#1}
	\end{figure*}
}

\newcommand{\picTwelve}[2]{
	\begin{figure*}[htb]
		\begin{subfigure}[c]{0.09\textwidth}		
			\includegraphics[width=\textwidth]{#1-00}
		\end{subfigure}
		\begin{subfigure}[c]{0.09\textwidth}		
			\includegraphics[width=\textwidth]{#1-01}
		\end{subfigure}
		\begin{subfigure}[c]{0.09\textwidth}		
			\includegraphics[width=\textwidth]{#1-02}
		\end{subfigure}
		\begin{subfigure}[c]{0.09\textwidth}		
			\includegraphics[width=\textwidth]{#1-03}
		\end{subfigure}			
		\begin{subfigure}[c]{0.09\textwidth}		
			\includegraphics[width=\textwidth]{#1-04}
		\end{subfigure}
		\begin{subfigure}[c]{0.09\textwidth}		
			\includegraphics[width=\textwidth]{#1-05}
		\end{subfigure}
		\begin{subfigure}[c]{0.09\textwidth}		
			\includegraphics[width=\textwidth]{#1-06}
		\end{subfigure}
		\begin{subfigure}[c]{0.09\textwidth}		
			\includegraphics[width=\textwidth]{#1-07}
		\end{subfigure}
		\begin{subfigure}[c]{0.09\textwidth}		
			\includegraphics[width=\textwidth]{#1-08}
		\end{subfigure}
		\begin{subfigure}[c]{0.09\textwidth}		
			\includegraphics[width=\textwidth]{#1-09}
		\end{subfigure}

		\begin{subfigure}[c]{0.09\textwidth}		
			\includegraphics[width=\textwidth]{#1-10}
		\end{subfigure}
		\begin{subfigure}[c]{0.09\textwidth}		
			\includegraphics[width=\textwidth]{#1-11}
		\end{subfigure}
		\begin{subfigure}[c]{0.09\textwidth}		
			\includegraphics[width=\textwidth]{#1-12}
		\end{subfigure}
		\begin{subfigure}[c]{0.09\textwidth}		
			\includegraphics[width=\textwidth]{#1-13}
		\end{subfigure}			
		\begin{subfigure}[c]{0.09\textwidth}		
			\includegraphics[width=\textwidth]{#1-14}
		\end{subfigure}
		\begin{subfigure}[c]{0.09\textwidth}		
			\includegraphics[width=\textwidth]{#1-15}
		\end{subfigure}
		\begin{subfigure}[c]{0.09\textwidth}		
			\includegraphics[width=\textwidth]{#1-16}
		\end{subfigure}
		\begin{subfigure}[c]{0.09\textwidth}		
			\includegraphics[width=\textwidth]{#1-17}
		\end{subfigure}
		\begin{subfigure}[c]{0.09\textwidth}		
			\includegraphics[width=\textwidth]{#1-18}
		\end{subfigure}
		\begin{subfigure}[c]{0.09\textwidth}		
			\includegraphics[width=\textwidth]{#1-19}
		\end{subfigure}

		\caption{#2}
		\label{fig:#1}
		
	\end{figure*}
}

\newcommand{\tblPretextTask}{
\begin{table}

	\centering
	\resizebox{0.85\linewidth}{!}{%
		
		\begin{tabular}{l c c c c }
			\toprule
			& & \multicolumn{2}{c}{Accuracy} &    \\ 
			\cmidrule(r){3-4}
			Method  & r & Overcluster & Normal & Runtime \\ 
			\midrule				
			IIC \cite{iic} & - &   63.1 & & -- \\
			
			\midrule 
			
			IIC* & - & 59.31 & & 114 h \\
			FOC  & 0 &  57.28 & 48.10 & \textbf{10 h}\\
			FOC  & 0.5 &  59.24 & \textbf{52.96} & 13 h\\
			FOC & 1& \textbf{64.44} & 50.92  & 89 h \\
			
			\midrule
			
			FOC$^\dagger$ & 1 & 72.06 & 69.83 & 81 h \\
			
			\bottomrule
		\end{tabular}
		
	}
	
	\caption{Pretext task on STL-10 -- 
		$r$ indicates the restriction of unlabeled data.
		The runtime is the training time in hours on the same hardware.
		The best results are marked bold.
		A horizontal line indicates different comparisons. 				
		Legend: * Original authors code used. 
		$^\dagger$ Uses supervised augmentations and can be seen as upper bound estimation.
	}
	\label{tbl:unsupervised-pretext-stl10}
	\vspace{-4mm}
\end{table}
}

\newcommand{\tblComparisonSTL}{
\begin{table}

		\centering
		
		\resizebox{\linewidth}{!}{%
			\begin{tabular}{l l  c c}
				\toprule
				& &  \multicolumn{2}{c}{Accuracy}     \\ 
				\cmidrule(r){3-4}
				Method & Network &  Overcluster & Normal  \\ 
				\midrule		
				ADC \cite{adc}  & ResNet &  &  77.0 \cite{iic} \\ 	
				SCAN$^\star$ \cite{scan} & ResNet18 & & 80.9 \\			
				IIC* \cite{iic} & ResNet34 &  & 85.76 \\	
				CMC$^\dagger$  \cite{cmc} & AlexNet &  & 86.88 \\ 	
				FOC (Ours) & ResNet34 &  85.54 & 86.98 \\
				IIC$^\dagger$ \cite{iic} & ResNet34 &  & 88.8 \\
				MixMatch$^\ddagger$ \cite{mixmatch}  & Wide ResNet28 &  & 89.82 \\
				MixMatch \cite{mixmatch} & Wide ResNet28 &  & 94.41 \\
				FixMatch$^\ddagger$ \cite{fixmatch} & Wide ResNet28 &  & 94.83 \\
				
				\bottomrule
			\end{tabular}
		}

			\caption{Comparison to state-of-the-art on STl-10  -- The results are reported in the original paper or the link is given behind the value.  Legend: * Original authors code without MLP for fine-tuning.
			$^\dagger$ A MLP used for fine-tuning.
			$^\ddagger$ Used only 1000 labels instead of 5000.
			$^\star$ Unsupervised method.
		}
		\label{tbl:comparisonSTL}
\end{table}
}

\newcommand{\tblComparisonPlankton}{
\begin{table}
		\centering

		\resizebox{\linewidth}{!}{%
			\begin{tabular}{l  c c c c}
				\toprule
				Method	& \# Classes &  F1-Score & Accuracy\\
				
				\midrule

				SCAN$^\star$ \cite{scan} & 10 & 38.34 & 62.54  \\	
				SCAN$^\star$ \cite{scan} & 60 & 38.15 & 62.16   \\		
				IIC \cite{iic} & 10 & 66.63 &  74.25   \\
				IIC$^\dagger$ \cite{iic} & 10 & 69.92 &  78.71   \\				
				Mean-Teacher \cite{mean-teacher}  & 10 & 74.51 & 81.06   \\
				Pi \cite{temporal-ensembling} & 10 & 74.03 & 81.09  \\
				Pseudo-label \cite{pseudolabel} & 10  & 75.14 & 81.12 \\
				FixMatch \cite{fixmatch} & 10 & 75.97 & \textbf{83.14} \\		
				FOC (Ours) &  10 & 67.23  &  79.00  \\
				FOC (Ours) &  60 & \textbf{77.60} &  \textbf{83.11} &   \\

				\bottomrule
			\end{tabular}	
		}
		
			\caption{Comparison to state-of-the-art on plankton dataset -- 
			The results were calculated on the unlabeled data which include the fuzzy labels.
			The number of classes indicates the usage of overclustering (60 classes).	
			The accuracy is provided for comparison but can be misleading due to class imbalance.
			All results within a one percent margin of the best result are marked bold. 
			Legend:
			$^\dagger$ A MLP used for fine-tuning.
			 $^\star$ Unsupervised method.
		}
		\label{tbl:comparisonPlankton}
		
		\vspace{-4mm}
		
\end{table}
}			
		
\newcommand{\tblConsistency}{	
\begin{table}

		\centering
		
		\resizebox{\linewidth}{!}{%
		\begin{tabular}{l c c c c }
			\toprule
	
			& \multicolumn{2}{c}{all data}  & 	\multicolumn{2}{c}{ignore class no-fit} \\ 
			
			\cmidrule(r){2-3} \cmidrule(r){4-5}
			Method  & overall & per cluster & overall & per cluster \\
			\midrule

			FixMatch \cite{fixmatch} &  82.56 & 78.78 $\pm$ 28.22 & 77.11 & 69.61 $\pm$ 29.41 \\
			FOC (Ours) &\textbf{87.80} & \textbf{79.66 $\pm$ 18.88} & \textbf{86.31} & \textbf{86.41 $\pm$ 13.68}\\
			
			\bottomrule
		\end{tabular}
		}
		
		\caption{Consistency comparison on plankton dataset --
		    The consistency is rated by experts over the complete data and a subset without the mixed class no-fit.
		    An image is consistent with the cluster if it is rated by an expert as similar with the majority of the cluster.
		    We provide the percentage of consist data overall and as average with standard deviation per cluster.
		    Further information about the calculations are provided in \autoref{subsec:consistency}.
		    }
		 \label{tbl:consistency} 
		 
		 \vspace{-4mm}
		   
\end{table}

}

\newcommand{\tblAblation}{	
\begin{table}
	\centering
	
	\begin{minipage}{\linewidth}
		\resizebox{\linewidth}{!}{%
		\begin{tabular}{l c c c c c c}
			\toprule
			& & & & & \multicolumn{2}{c}{Accuracy}     \\ 
			\cmidrule(r){6-7}
			Method & Pre & MI & \ceinv &  Weight &    Overcluster & Normal \\
			\midrule
			FOC  &  & X& & -- &   75.71 & 76.46 \\	
			IIC* \cite{iic} &  X  & & & -- &   & 85.76 \\				
			FOC  &  X & X &  & -- & 74.19 & 86.05 \\	
			FOC &  X & X& X & -- &  82.69& 86.49 \\		
			FOC  &  X & X & X & C &  \textbf{85.54} & \textbf{86.98}\\
			FOC  & X & X&X & I & 83.70 & 85.25 \\
			
			\bottomrule
		\end{tabular}
	}
	\subcaption{STL-10}
	
	\end{minipage}%
	\hfill
	\begin{minipage}{\linewidth}
		\resizebox{\linewidth}{!}{%
			\begin{tabular}{l c c c c c c c c}
				\toprule
				& & & & &  \multicolumn{2}{c}{Accuracy} &   \multicolumn{2}{c}{F1-Score}     \\ 				
				\cmidrule(r){6-7}  \cmidrule(r){8-9} 
				Method & Pre & MI & \ceinv &  Weight &    Overcluster & Normal &  Overcluster & Normal  \\ 
				\midrule
				
				IIC \cite{iic}  & X &  &  & -- &  -- & 74.25 & -- & \textbf{66.63} \\
				IIC$^\dagger$ \cite{iic} & X &  &  & -- &  -- &  78.71 &  -- & \textbf{69.92}  \\		
				
				\midrule
				
				FOC & & &  & C & 51.17 & 59.31 & 31.45 & 38.70 \\
				FOC & & X & & C & 58.24 & \textbf{79.28} & 32.27 & 60.66 \\
				FOC & & X & X & C  & 78.66  & 76.68 & 71.70 & 64.23 \\
				
				\midrule
				FOC & X & &  & C & 55.64 & 73.94 & 27.64 & 59.84 \\
				FOC& X & X& & C & 80.63 & 75.12 & 70.37 & 56.67 \\
				FOC & X & X & X & C  & \textbf{82.25} & 77.64 & 70.57 & 58.12 \\
				
				\midrule
				
				FOC & & & & I & 55.21 & 74.69 & 29.57 & 54.92 \\		
				FOC & & X & & I & 80.43 & 78.42 & \textbf{72.34} & 64.80 \\
				FOC &  & X& X & I & \textbf{81.48} & 78.81 & \textbf{73.52} & 64.85 \\
				
				\midrule
				FOC & X & &  & I & 59.87 & 74.73 & 39.54 & 55.54 \\
				FOC & X & X & & I & 80.63 & \textbf{80.48} & 70.66 &  66.47 \\
				FOC & X & X & X & I &   \textbf{83.11}  & \textbf{79.00} & \textbf{77.60}  & \textbf{67.23}  \\
				
				\bottomrule
			\end{tabular}
		}
		
		\subcaption{plankton dataset}

	\end{minipage}%
	
	 \caption{Ablation study on STL-10 and plankton dataset --
		The second to fourth column indicate if a pretext task as pretraining, the mutual information loss and our novel loss \ceinv was used respectively.
		The fifth column indicates the weights used for initialization.
		C stands for Cifar-20 and I for ImageNet.
		If no weight initilization is given, sobel filtered images are used as input.
		The best results are marked bold for STL-10 and the best three for plankton dataset.
		Legend: 
		    * Original authors code 
			$^\dagger$ A MLP used for fine-tuning.
		}	
			\label{tbl:Ablation}	
			
	\vspace{-4mm}
\end{table}
		    
}

\newcommand{\ceinv}{CE$^{-1}$}

\title{Beyond Cats and Dogs: Semi-supervised Classification of fuzzy labels with overclustering}

\author{
	Lars Schmarje\textsuperscript{\rm 1}
	\thanks{ Corresponding Author}
	\qquad Johannes Brünger\textsuperscript{\rm1}
	\qquad Monty Santarossa\textsuperscript{\rm1}\\
	Simon-Martin Schröder\textsuperscript{\rm1}
	\qquad Rainer Kiko\textsuperscript{\rm 2}
	\qquad Reinhard Koch\textsuperscript{\rm1}\\
	{\textsuperscript{\rm 1} Multimedia Information Processing Group, Kiel University, Germany}  \\
	{\textsuperscript{\rm 2} Laboratoire d'Océanographie de Villefranche, Sorbonne Université, France} \\
	{\tt \small{\{las, jbr, msa, sms, rk\}@informatik.uni-kiel.de}} \qquad
	{\tt \small{rainer.kiko@obs-vlfr.fr}}
}

\begin{document}
	
	\maketitle
	
	\begin{abstract}
		A long-standing issue with deep learning is the need for large and consistently labeled datasets.
		Although the current research in semi-supervised learning can decrease the required amount of annotated data by a factor of 10 or even more, this line of research still uses distinct classes like cats and dogs.
		However, in the real-world we often encounter problems where different experts have different opinions, thus producing fuzzy labels.
		We propose a novel framework for handling semi-supervised classifications of such fuzzy labels.
		Our framework is based on the idea of overclustering to detect substructures in these fuzzy labels.
		We propose a novel loss to improve the overclustering capability of our framework and show on the common image classification dataset STL-10 that it is faster and has better overclustering performance than previous work.
		On a real-world plankton dataset, we illustrate the benefit of overclustering for fuzzy labels and show that we beat previous state-of-the-art semi-supervised methods.
		Moreover, we acquire 5 to 10\% more consistent predictions of substructures. \footnote{The source code, the datasets and supplementary material will be released soon. Please contact the corresponding author for any questions.}
	\end{abstract}
	
	\section{Motivation}
	
	\picSmall{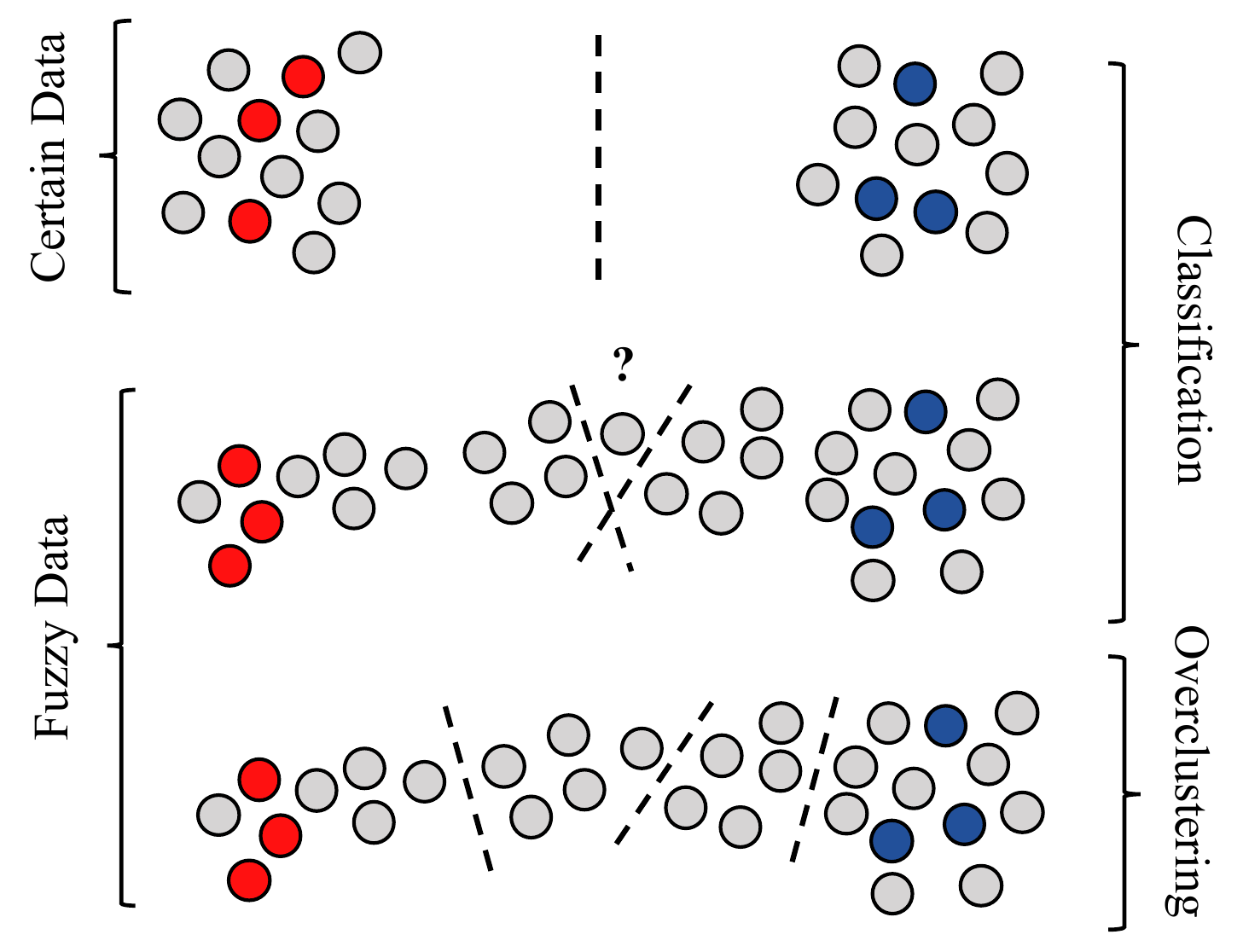}{Illustration of fuzzy data and overclustering - 
		The grey dots represent unlabeled data and the colored dots labeled data from different classes.
		The dashed lines represent decision boundaries.
		For certain data, a clear separation of the different classes with one decision boundary is possible and both classes contain the same amount of data (top).
		For fuzzy data determining a decision boundary is difficult because of fuzzy datapoints between the classes (middle).
		If you overcluster the data, you get smaller but more consistent substructures in the fuzzy data (bottom).}{0.95}
			
	Over the past years, we have seen the successful application of deep learning to many computer vision problems.
	We can classify images of cats and dogs easily while state-of-the-art semi-supervised methods can decrease the needed amount of annotated data by a factor of 10 or even more \cite{remixmatch,S4L,simclr}.
	
	However, if we go beyond cats and dogs into the classification of different breeds of dogs and even their crossbreeds, we will have most likely many experts with many different opinions. 
	This issue is also known as intra- and inter-observer variability \cite{noisy-labels-comparison} and is common in biological and medical application fields~\cite{tailception,schmarje2019,eye-fuzzy,cancergrading,medicinecrowdsource,planktonUncertain}.
	Due to image quality issues, even the distinction of written letters can become complicated~\cite{ocr-example}.
	Also in other computer vision tasks like image segmentation, a fine and consistent annotation of images is especially hard to achieve~\cite{cityscapes,inter-observer-segmentation}.
	In this \emph{fuzzy data}, the real label can only be obtained with strict guidelines, finding consensus or averaging multiple annotations \cite{staple}.
	All these approaches have in common that they are time-consuming and therefore expensive.
	
	The underlying issues of fuzzy data like skewed class distributions and noisy labels are addressed in recent methods \cite{remixmatch,divide-mix} and literature reviews \cite{surveynoise,surveynoise2}.
	However, we argue that fuzzy data should not be interpreted just as label noise.
	If an object is classified by one half of the experts as A and by the other half as B, this object should not be arbitrarily classified into class A or B (see the middle image in \autoref{fig:idea}).
	In agreement with domain experts, we argue that identifying substructures in the fuzzy data is more valuable than forcing them to fit into known classes because substructure detection allows us to treat each substructure according to the target problem.
	The different treatments could be sorting them into an existing or a new class, ignore them during evaluation or reweighting their impact.
	The ability to handle fuzzy data and identify substructures is the main contribution of this work (see the bottom image in \autoref{fig:idea}).
			
	In this paper, we propose the first framework for Fuzzy Overclustering (FOC) in order to handle these \emph{fuzzy labels}.
	We call them fuzzy labels because it is difficult to distinguish the classes for such labels.
	The correct label might even be a combination of different classes, for example for crossbreeds.
	Our idea is to rephrase the handling of fuzzy labels as a semi-supervised learning problem by using a small set of certainly labeled images and a large number of fuzzy images that are treated as unlabeled.
	This approach allows us to use the idea of overclustering from semi-supervised literature \cite{iic,deep-cluster} and apply it to fuzzy data.
	The difference to previous work is that we use overclustering not only to improve classification accuracy on the labeled data but improve the clustering and therefore the identification of substructures of fuzzy data.
	We show that overclustering allows us to cluster the fuzzy images in a more meaningful way by finding substructures and therefore allowing experts to analyze fuzzy details more consistently in the future.
	
	Our Framework FOC uses a novel loss to improve the overclustering performance: \emph{Inverse Cross-entropy}. 
	We will show that FOC  beats state-of-the-art semi-supervised methods on fuzzy data.
	The framework can restrict the used data per epoch to improve the training time by a factor of 9 in comparison to previous methods.
	Additionally, we achieve 5 to 10\% more consistent predictions which is important for later expert analysis.

	Our key contributions are:
	\begin{itemize}
		\item
		We propose a novel framework for handling fuzzy labels with a semi-supervised approach.
		This framework uses overclustering to find substructures in fuzzy data.
		\item We show that overclustering is a powerful tool for analyzing fuzzy data and propose a novel loss, inverse cross-entropy, which improves the overclustering quality in semi-supervised learning. 
		\item We show that our framework can cut the training time on STL-10 from days to hours compared to previous methods.
		We show on a real-world plankton dataset with fuzzy labels and a skewed class distribution that we outperform previous state-of-the-art semi-supervised methods like FixMatch \cite{fixmatch}.
		Moreover, we achieve 5 to 10\% more self-consistent predictions on fuzzy labels.
		
	\end{itemize}

	\section{Related Work}
	\label{subsec:related}
	
	Semi-supervised learning in general and handling noisy data are two ideas to handle the proposed problem of fuzzy labels.
	In the following, we will describe these two topics in detail.	

	\paragraph{Semi-Supervised Learning}
	
	Semi-supervised learning~\cite{semi-supervised-learning} aims at training a network with a combination of labeled and unlabeled data.
	The goal is to decrease the required amount of labeled data without a significant impact on the final accuracy.
	Many state-of-the-art methods use a combination of different ideas to leverage unlabeled data.
	A common strategy is to define a pretext task like image rotation prediction~\cite{self-rotation} or jigsaw puzzle solving~\cite{self-jigsaw, self-jigsaw++}.
	This task can either be used to pretrain the network or as an additional loss~\cite{S4L,remixmatch}.
	Moreover, consistency regularization is often used to either encourage consistent~\cite{mean-teacher, temporal-ensembling, S4L} or highly confident~\cite{entropy-min} predictions on unlabeled data.
	A broad overview of current trends, ideas and methods in semi-, self- and unsupervised learning is available in~\cite{survey}.
	
	State-of-the-art semi-supervised methods~\cite{S4L,remixmatch,fixmatch,enaet,simclr} achieve almost the same results as supervised methods with only 10\% of the labeled data.
	These results are commonly reported on well-established image classification databases like CIFAR-10 \cite{cifar}, STL-10 \cite{stl-10} or ImageNet \cite{imagenet}.
	However, these datasets do not contain fuzzy labels and therefore many semi-supervised methods can only make an arbitrary decision boundary (see middle in \autoref{fig:idea}).
	This leads to ambiguous predictions for intermediate objects and does not allow the identification of substructures.
	We developed our method on STL-10 but we will show that it can be applied to a real-world plankton dataset with fuzzy labels without hyperparameter tuning.
	
	\picTwo{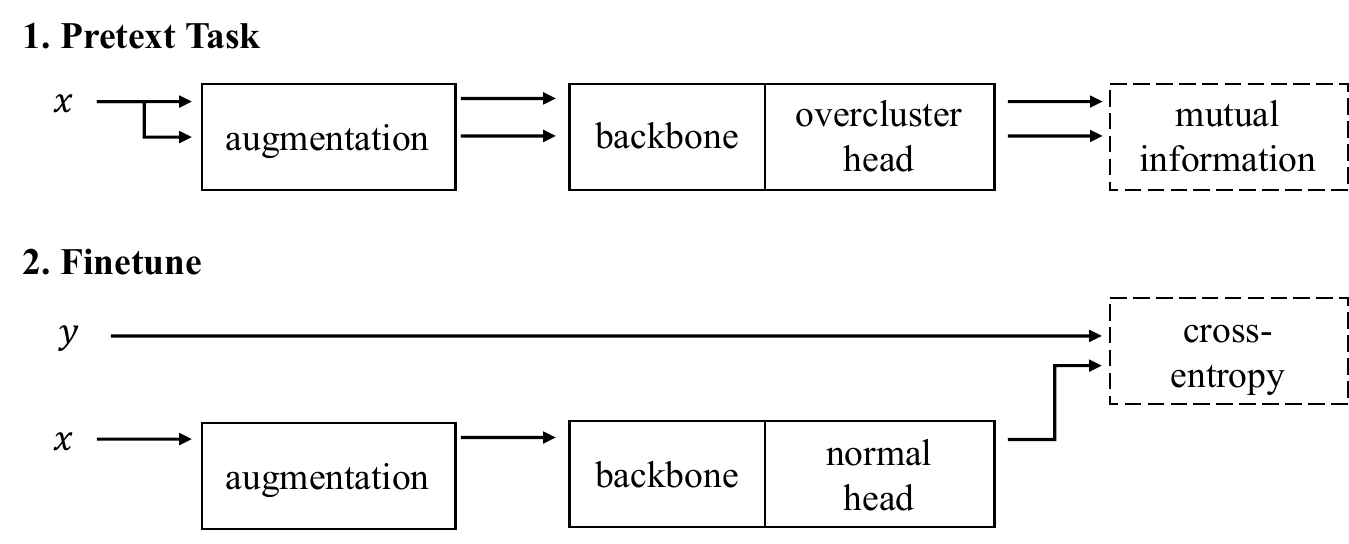}{IIC}{FOC}{Overview of the two subsequent tasks of IIC and our unified framework FOC and  for semi-supervised classification --
		The input image is $x$ and the corresponding label is $y$.
		The arrows indicate the usage of image or label information.
		Parallel arrows represent the independent copy of the information.
		The usage of the label for the augmentations is described in \autoref{subsec:aug}.
		The red arrow stands for an inverse example image $x'$ with a different label than $y$.
		The output of the normal and the overclustering head have different dimensionalities.
		The normal head has as many outputs as ground-truth classes exist ($k_{GT}$) while the overclustering head has $k$ outputs with  $k > k_{GT}$.
		The dashed boxes on the right side show the used loss functions.
		More information about the losses mutual information and inverse cross-entropy can be found in \autoref{subsec:mi} and \autoref{subsec:ceinv} respectively.
	}

	\paragraph{Handling Noisy or Fuzzy Data}
	
	Noisy labels are a common data quality issue and are discussed in the literature \cite{noisy-labels-comparison,surveynoise,surveynoise2}.
	Fuzzy labels are known as a special case of label noise that exist ''due to subjectiveness of the task for human experts or the lack of experience in annotator[s]'' \cite{surveynoise}.
	In contrast to us, most methods \cite{self-noisy,temporal-ensembling,divide-mix} and literature surveys \cite{noisy-labels-comparison,surveynoise,surveynoise2} interpret fuzzy labels as corrupted labels.
	We argue that fuzzy labels are ambiguous in nature and that it is important to discover the substructures for real-world data handling \cite{tailception,schmarje2019,eye-fuzzy,cancergrading,medicinecrowdsource,planktonUncertain}.
	
	Geng proposed to learn the label distribution to handle fuzzy data \cite{labeldistribution} and the idea was extended to the application of real-world images \cite{deep-learn-label-distribution}.
	However, these methods are not semi-supervised and therefore depend on large labeled datasets.
	A variety of methods was proposed to handle fuzzy data in a semi-supervised learning approach \cite{fuzzy-semi-supervised,fuzzymeta,semi-supervised-soft}.
	These methods use lower-dimensional features spaces in contrast to images as input.
	Liu et al. proposed a semi-supervised learning approach in the context of photo shot-type classification \cite{fuzzyPhotos}.
	They use independent predictions of multiple networks as pseudo-labels for the estimation of the label distribution.
	We argue that the true label distribution is difficult to approximate and therefore also to evaluate.
	We do not learn the label distribution but use overclustering to identify substructures. 
	The substructures can be evaluated by experts for their consistency and therefore their quality.

	\section{Fuzzy Overclustering (FOC)}
	\label{sec:methods}
	
	Our framework Fuzzy Overclustering (FOC) is an extension of Invariant Information Clustering (IIC) \cite{iic} with a focus on overclustering.
	The are many small differences between the methods but the major differences are the novel loss \ceinv (\autoref{subsec:ceinv}), supervised augmentations (\autoref{subsec:aug}), restricted unlabeled data (\autoref{subsec:restricted}) and the warm up (\autoref{subsec:warmup}).
	A graphical comparison of the methods is given in \autoref{fig:pipeline}.
	
	Let $X$ be an arbitrary set of images and be composed of unlabeled $X_u$ and labeled $X_l$ data.
	For all images $x \in X_l$ we have a label $y \in L$.
	We define $k_{GT}$ as the number of different labels in $L$. 	
	A neural network $\Phi$ is composed of a backbone like ResNet50 \cite{resnet} and a linear prediction layer.
	Following \cite{iic}, we call this linear predictor \emph{head} and use either normal heads or overclustering heads.
	The output of a network with a normal head is $\Phi_n(x) \in \{1, ..., k_{GT}\}$.
	The output of a network with an overclustering head is  $\Phi_o(x) \in \{1, ..., k\}$ with more clusters than ground-truth classes ($k > k_{GT}$).
	
	The central idea of IIC is that augmented versions $x_1 = g_1(x),x_2 = g_2(x)$ of the same image $x \in X$ should be placed in the same cluster,  where $g_1$ and $g_2$ are random augmentation that can include for example translation, rotation or color change.
	Ji et al. propose to maximize the mutual information $\mathcal{L}_{MI}(\Phi(x_1),\Phi(x_2))$ between the representations $\Phi(x_1)$ and $\Phi(x_2)$ produced by a neural network $\Phi$ in order to produce these clusterings \cite{iic}.
	This maximization is described in detail in \autoref{subsec:mi}.
	IIC is trained for semi-supervised classification in two stages: Pretext task and Fine-tune.
	In the pretext task, IIC uses a modified ResNet34 backbone with overclustering head $\Phi_o$ and is trained with the loss $\mathcal{L}_{MI}$ and all data $X$ without the labels (see the top image in \autoref{fig:pipeline-0}).
	IIC fine-tunes a network with normal head $\Phi_n$ and shared backbone to $\Phi_o$ on the labeled data $X_l$ with cross-entropy (see the bottom image in \autoref{fig:pipeline-1}).
	Ji et al. use sobel-filtered images as input to train the network on structure information rather than color distributions.
	
	In general, we keep the two stages from IIC.
	We will see in \autoref{subsec:warmup} why we can interpret FOC also as a method with only one stage.
	In the pretext task, we use a network with normal and overclustering heads $\Phi_{on}$ and train on all data $X$.
	We use the loss $\mathcal{L}_{MI}$ but reduce the influence of the unlabeled data on the training time by a manifold as we will show in \autoref{tbl:unsupervised-pretext-stl10}.
	The details are described in \autoref{subsec:restricted}.
	During fine-tuning, we continue to use a network with normal and overclustering heads $\Phi_{on}$, train on all data $X$ and use the loss $\mathcal{L}_{MI}$.
	Additionally, we use the labels of the labeled dataset $X_l$ in three ways.
	Firstly, we calculate the cross-entropy on the normal head for $\Phi_n(x_1)$ with its label $y$.
	Secondly, the label is used to create supervised augmentations of $x$.
	The details are described in \autoref{subsec:aug}.
	Thirdly, we choose an image $x'$ with a different label $y'$ than the input $x$ and its label $y$.
	We use an augmented version of this image as third input $x_3 = g_3(x')$.
	This third input is used to calculate our novel loss inverse cross-entropy  (\ceinv) on the overclustering to improve the overclustering performance.
	The details of \ceinv are described in \autoref{subsec:ceinv}.
	
	To sum up, for both heads the loss is different but can be written as the weighted sum of an unsupervised and a supervised loss as follows:
	\begin{equation}
	\label{eq:loss}
	\mathcal{L} = \lambda_s \mathcal{L}_s + \lambda_u \mathcal{L}_u
	\end{equation}
	For both heads $\mathcal{L}_u$ is the mutual information loss $\mathcal{L}_{MI}$. 
	$\mathcal{L}_s$ is cross-entropy ($\mathcal{L}_{CE}$) for the normal head  and our novel \ceinv loss ($\mathcal{L}_{CE^{-1}}$) for the overclustering head.
	An illustration of the complete pipeline is given in \autoref{fig:pipeline-1}.
	We initialize our backbones with pretrained weights and can therefore directly use RGB images as input. 
	For further implementation details see \autoref{subsec:implementation}.
	
	Our framework FOC can also be used to perform standard unsupervised clustering as done by IIC.
	The details about unsupervised clustering and a comparison to IIC is given in the supplementary. 

	\subsection{Mutual Information (MI)}
	\label{subsec:mi}
	
	We want to maximize the mutual information between two output predictions $\Phi(x_1), \Phi(x_2)$ with  $x_1, x_2$ images which should belong to the same cluster and $\Phi :  X \rightarrow [0,1]^k$ a neural network with $k$ output dimensions.
	We can interpret $\Phi(x)$ as the distribution of a discrete random variable $z$ given by $P(z = c | x) = \Phi_c(x)$ for $c \in \{1, \dots, k\}$ with $\Phi_c(x)$ the c-th output of the neural network.
	With $z,z'$ such random variables we need the joint probability distribution for $P_{cc'} = P(z=c, z'=c')$ for the calculation of the mutual information $I(z,z')$.
	Ji et al. propose to approximate the matrix $P$ with the entry $P_{cc'}$ at row $c$ and column $c'$ by averaging over the multiplied output distributions in a batch of size $n$ \cite{iic}.
	Symmetry of $P$ is enforced as follows:
	\begin{equation}
	P = \frac{Q + Q^T}{2} \mbox{ with } Q = \frac{1}{n} \sum_{i=1}^{n} \Phi(x_i) \cdot \Phi(x_i')^T     
	\end{equation}
	With the marginals $P_c = P_{c'} = P(z = c)$ given as sums over the rows or columns we can maximize our objective $I(z,z')$:
	\begin{equation}
	I(z,z') = \sum_{c=1}^{k}\sum_{c'=1}^{k} P_{cc'} \cdot ln \frac{P_{cc'}}{P_{c} \cdot P_{c'}}
	\end{equation}
	
	\subsection{Inverse Cross-Entropy (\ceinv)}
	\label{subsec:ceinv}
	
	For normal heads, we can use cross-entropy (CE) to penalize the divergence between our prediction and the label.
	We can not use CE directly for the overclustering heads since we have more clusters than labels and no predefined mapping between the two.
	However, we know that the inputs $x_1 / x_2$ and $x_3$ should not belong to the same cluster. 
	Therefore, our goal with \ceinv is to define a loss that pushes their output distributions (e.g $\Phi(x_1)$ and  $\Phi(x_3)$) apart from each other.
	
	Let us assume we could define a distribution that $\Phi(i_3)$ should not be.
	In short, an inverse distribution $\Phi(x_3)^{-1}$.
	If we had such a distribution we could use CE to penalize the divergence for example between $\Phi(x_1)$ and $\Phi(x_3)^{-1}$.
	
	One possible and easy solution for an inverse distribution is $\Phi(x_3)^{-1} = 1 - \Phi(i_3)$. 
	For a binary classification problem, $\Phi(x_3)^{-1}$ can even be interpreted as a probability distribution again. 
	This is not the case for a multi-class classification problem.
	We could use a function like softmax to cast $\Phi(x_3)^{-1}$ into a probability distribution but decided against it for three reasons.
	Firstly, we would penalize correct behavior. 
	For example in a three class problem with $\Phi_1(x_1) = 0.5 = \Phi_2(x_1)$ and $\Phi_3(x_3) = 1$ we only get $CE(\Phi(x_1), \Phi(x_3)^{-1}) = 0$ if $\Phi(x_3)^{-1}$ is not a probability distribution.
	Otherwise either $\Phi_1(x_3)^{-1}$ or $\Phi_2(x_3)^{-1}$ have to be real smaller than 1.
	Secondly, we are still minimizing the entropy of $\Phi(x_1)$ which leads to more confident predictions in semi-supervised learning \cite{entropy-min,S4L,UDA,vat,mixmatch,remixmatch}.
	The proof is given in the supplementary. 
	Thirdly, it is easier and in practice, it is not needed. 
	For the input $i = (x_1, x_2, x_3)$, we define the cross-entropy inverse loss $\mathcal{L}_{CE^{-1}}$ as
	\begin{equation}
	\begin{split}
	\mathcal{L}_{CE^{-1}}(i) & = 0.5 \cdot CE^{-1}(\Phi(x_1), \Phi(x_3)) \\
	& + 0.5 \cdot CE^{-1}(\Phi(x_2), \Phi(x_3)) \,\mbox{, with } \\
	CE^{-1}(p,q) & = - \sum_{c=1}^{k} p(c) \cdot ln(1-q(c))\,.
	\end{split}
	\end{equation}
	
	\subsection{Supervised Augmentations}
	\label{subsec:aug}
	
	In the unsupervised pretraining, we use the same image $x$ to create the two inputs $x_1 = g_1(x)$ and $x_2 = g_2(x)$ based on the augmentations $g_1$ and $g_2$. 
	Otherwise, without supervision, it is difficult to determine similar images. 
	However, if we have the label $y$ for $x$ we can use a secondary image $x' \in X_l$ with the same label.
	In this case we can create $x_2 = g_2(x')$ based on the different image.
	We call this \emph{supervised augmentation}.
	
	Supervised augmentations can easily be implemented and are an upper bound estimate for the performance of optimal augmentations in the pretext task training.
	During Fine-tuning, this performance boost is integrated into the model.

	\subsection{Restricted Unsupervised Data}
	\label{subsec:restricted}
	
	We will see in \autoref{sec:evaluation} that unlabeled data has a small impact on the results but drastically increases the runtime in most cases.
	The increased runtime is caused by the facts that we often have much more unlabeled data than labeled data and that a neural network runtime is normally linear in the number of samples it needs to process.
	For example, STL-10 consists of 95\% unlabeled data.
	However, unlabeled data is essential for our proposed framework and we can not just leave it out.
	We propose to restrict the unlabeled data to a fixed ratio $r$ in every batch and therefore the unlabeled data per epoch.
	The ratio $r$ is an upper bound for the unlabeled data and less unlabeled data can be used in each batch if it is not available.
	Detailed examples are given in the supplementary.
	
	It is important to notice that we restrict only the unlabeled data per batch/epoch.
	While for one epoch the network will not process all unlabeled data, over time all unlabeled data will be seen by the network.
	We argue that the impact on training time negatively outweighs the small benefit gained from all unlabeled data per epoch as we will show in \autoref{sec:evaluation}.
	
	\subsection{Warm up}
	\label{subsec:warmup}
	
	Up to this point, we described FOC to be trained in two stages like IIC. 
	In contrast to IIC, we use the same data and network for the pretext task and the fine-tuning.
	If we use FOC with $\lambda_s = 0$ and without supervised augmentations there is no difference to the pretext task model.
	Due to this fact, our framework can be used as a single stage training algorithm with a warm up phase based on the pretext task.
	During the evaluation, we will keep referencing the individual stages of FOC for a better comparison to IIC.
	
	\section{Evaluation}
	\label{sec:evaluation}
	
	We conducted experiments on the common image classification dataset STL-10 and a real-world plankton dataset.
	We show that our framework can be applied to both datasets without hyperparameter tuning between them.
	For both datasets we compare ourselves to previous work, make ablations and a rough runtime performance comparisons.
	Additional results like unsupervised clustering, ablations and further details are given in the supplementary material.
	
	While the issue of fuzzy labels is present in multiple datasets \cite{tailception,schmarje2019,eye-fuzzy,cancergrading,medicinecrowdsource,planktonUncertain}, they are not well suited for evaluations.
	If we want to quantify the performance on fuzzy labels, we need a dataset with very good ground-truth.
	This can only be achieved with multiple annotations and their aggregation. 
	For many datasets, it is not feasible to create multiple annotations for a large number of images.
	We will see below that the proposed Plankton dataset is quite unique in this regard due to the help of citizen scientists.	
	
	\subsection{Datasets}
	
	\paragraph{STL-10}
	
	STL-10 is a common semi-supervised image classification dataset \cite{stl-10}.
	It consists of 5,000 training samples and 8,000 validation samples depicting every-day objects.
	Additionally, 100,000 unlabeled images are provided.
	It is important to notice that these unlabeled images may belong to the same or different classes than the training images.
	All images are colored and have a size of 96x96 pixels.
	These images were generated from the ImageNet dataset~\cite{imagenet}.
	In contrast to the plankton dataset, no labels are provided for the unlabeled data.
	
	\paragraph{Plankton}
	
	The plankton dataset contains diverse grey level images of marine planktonic organisms.
	The images were captured with an Underwater Vision Profiler~5~\cite{uvp} and are hosted on EcoTaxa~\cite{ecotaxa}.
	In the citizen science project PlanktonID\footnote{\url{https://planktonid.geomar.de/en}}, each sample was classified multiple times by citizen scientists. 
	The dataset consists of  12,280 images in originally 26 classes.
	We merged minor and similar classes so that we ended up with 10 classes.
	The class no-fit represents a mixture of left-over classes.
	After this process, a class imbalance is still present with the smallest class containing about 4.16\% and the largest class 30.37\% of all samples.
	We use the mean over all annotations as the reference annotation.
	The citizen scientists agree on most images completely. 
	We call these images \emph{certain}.
	However, about 30\% of the data has as least one disagreeing annotation.
	We call these images \emph{fuzzy}.
	The fuzzy images are used only as unlabeled data.
	More details about the mapping process, the number of used samples and graphical illustrations are given in the supplementary.
	
	\subsection{Implementation Details}
	\label{subsec:implementation}
	
	As a backbone for our framework, we used either a ResNet34 variant proposed by \cite{iic} or a standard ResNet50v2~\cite{resnet}.
	The heads are single fully connected layers with a softmax activation function.
	Following \cite{iic}, we use five randomly initialized copies for each type of head and repeat images per batch three times for more stable training.
	We alternated between training the different types of heads.
	The inputs are either sobel-filtered images or color images and are fed in parallel through the system.
	A weight initialization is needed for the color images.
	For the ResNet34 backbone, we use CIFAR20 weights and for the ResNet50v2 backbone ImageNet~\cite{imagenet} weights. 
	CIFAR-20 refers to the 20 superclasses in CIFAR-100~\cite{cifar}.
	
	We train the framework with Adam and an initial learning rate of 1e-4 for 500 epochs.
	When switching from the pretext task to fine-tuning, we train only the heads for 100 epochs with a learning rate of 1e-3. The number of outputs for the overclustering head should be about 5 to 10 times the number of classes.
	The exact number is not crucial because it is only an upper bound for the framework. 
	We use 70 for STL-10 and 60 for the plankton dataset.
	We use $\lambda_s = 1 = \lambda_u$ and an unlabeled data restriction of $r = 0.5$ if not stated otherwise.
	A more detailed description, the used hardware and hyperparameters are given in the supplementary.

	\subsection{Metrics}
	
	We use different metrics to measure the performance in different stages.
	In the pretext task, we follow standard literature \cite{dac,iic,adc} and calculate the best accuracy for all one-to-one permutations between clusters and classes for the normal head.
	For the overclustering head, we use the ground-truth labels to calculate a mapping between clusters and classes on the training data before we calculate the accuracy of the validation data.
	We only report the result for the head with the  best validation accuracy.
	
	In contrast to STL-10, we report the unlabeled data scores for the plankton dataset because we are interested in the results on fuzzy data.
	On the unlabeled plankton data, we calculate the mapping based on the unlabeled data because we expect human experts to be involved in this process for the identification of substructures.
	We calculate the macro F1-score as the accuracy can be misleading due to the skewed class distribution.
	We report the best results based on the best validation accuracy for the unlabeled data.

	\subsection{Experiments}
	
	\subsubsection{STL-10}
	
	In \autoref{tbl:comparisonSTL}, we present the comparison to current state-of-the-art semi-supervised methods on STL-10.
	We see that FOC reaches a performance of about 87\% which is an average performance. 
	It is slightly better than IIC with around 86\% if the architecture is the same and is not using a multilayer perceptron (MLP) instead of a single layer for fine-tuning.
	However, FOC is not able to reach the performance of MixMatch or even FixMatch. 
	FixMatch outperforms FOC by a clear margin of nearly 8\% while using a fifth of the labels.
	It is important to notice that only IIC without MLP shares the same network as FOC.
	Kolesnikov et al.  showed that slight changes in the architecture can influence the performance significantly \cite{revisiting-self}. 
	We see for example the performance gain of IIC of about 3\% with an MLP.	
	This performance is expected as FOC does not focus like the others on classifying certain but fuzzy data.

	\tblComparisonSTL
	
	\subsubsection{Plankton dataset}	
	
    We compare the state-of-the-art methods on unlabeled data of the plankton dataset which include fuzzy labels in \autoref{tbl:comparisonPlankton}.
    We see that FOC with overclustering outperforms all other methods by at least nearly 3\%.
	The unsupervised method SCAN is separated from all other semi-supervised methods by around 30-40\% in the F1-Score. 
	This indicates the importance of some labeled examples if the classes are not easily separable.
	IIC performs similar to our method without overclustering but is outperformed by the other semi-supervised methods.
	FixMatch reaches the best semi-supervised classification performance with an F1-Score of 76\% and an accuracy of around 83\% .
	Our Framework FOC with overclustering reaches an F1-Score of about 78\% and similar accuracy .
	We conclude that the overclustering head is more suitable for handling fuzzy real-world data as we assumed at the beginning.
	In accordance with previous work~\cite{morphocluster}, we need clustering to distinguish a high number of meaningful classes in similar or fuzzy data.

	\tblComparisonPlankton
	
	\picTwelve{qualitative}{Qualitative results for unlabeled data -- 
		The results in each row are from the same predicted cluster.
		The three most important fuzzy labels based on the citizen scientists' annotations are given below the image. 
		The last two items with the red box in each row show examples not matching the majority of the cluster.}
	
	\subsubsection{Consistency}
	\label{subsec:consistency}
	
	Up to this point, we analyzed classification metrics based on the 10 ground-truth classes.
	However, the ground-truth label can only be approximated for fuzzy data and the quality of substructures can not be evaluated because no ground-truth is available.
	We can judge the consistency of each image with its cluster with the help of experts as a quality measure.
	An image is consistent if an expert views it as visually similar to the majority of the cluster.
	The consistency is calculated by dividing the number of consistent images by all images. 
	The consistency over all classes or per class for FOC and FixMatch is given in \autoref{tbl:consistency}.
	A table with raw numbers per cluster is provided in the supplementary.
	Due to the mixture of different plankton entities in the class no-fit the visual similarity is not decidable.
	Only if we interpret similarity as the image is not similar to the other nine classes the similarity can be judged.
	Due to this, we provide a comparison based on all data and without the no-fit class.
	
	Based on the F1-Score, FixMatch and FOC perform similarly but if we look at the consistency we see that FOC is more than 5\% more consistent than FixMatch.
	If we exclude the class no-fit from the analysis, FOC reaches a consistency of around 86\% in comparison to  77\% from FixMatch.
	For both sets, our method FOC reaches a higher average consistency per cluster and lower standard deviation.
	This means the clusters produced by FOC are more relevant in practice because there are fewer low-quality clusters which can not be used.

	\tblConsistency

	\subsubsection{Qualitative Results}
	
	We illustrate the performance of our framework with an F1-score of nearly 78\% and consistency of around 88\% by some qualitative results in \autoref{fig:qualitative}.
	All images in a cluster are visually similar, even the probably wrongly assigned images (red box).
	For the images in the first row, the annotators are certain that the images belong to the same class.
	In the second row, annotators show a high uncertainty of assignment between the two variants of the same biological object.
	This illustrates the benefit of overclustering since visual similar items are in the same cluster even for fuzzy annotations.
	In a consensus process for the second row, experts could decide if the cluster should be the puff, tuft or a new borderline class.
	We provide more and randomly selected results in the supplementary.

	\tblPretextTask

	\subsection{Ablation Study}	
	
	\subsubsection{Pretext task on STL-10}
	
	On the pretext task on the STL-10 data, we investigate the impact of $r$ on the runtime and the benefit of supervised augmentations.
	Moreover, we compare the results to IIC.
	The results are shown in  \autoref{tbl:unsupervised-pretext-stl10}.
	The direct comparison between IIC and FOC ($r=1.0$) shows that we outmatch IIC by 1.34\%.
	The reported results of IIC used a higher number of clusters than ours.
	We conduct our own experiment with the original source code and an equal number of clusters.
	Our restricted FOC ($r=0.5$) performs almost equally to this better comparable version of IIC.
	Due to the restriction of unlabeled data, our method has a lower runtime by the factor of about 9.
	If we allow no unlabeled data during training ($r=0.0$), we can speed up the training even further but the accuracy is decreasing as well by 2\%.
	As a comparison, we added FOC with $r=1.0$ and supervised augmentations as an upper bound estimation.
	We can identify a clear gap of 8 to 20\% in the performance with this low form of supervision.	
	
	\vspace{-2mm}
	
	\subsubsection{Fine-tuning}
	
	In \autoref{tbl:Ablation} multiple ablations for STL-10 and the plankton dataset are given.
	Based on these tables, we illustrate the impact of the pretext task, the initialization and the usage of the MI and \ceinv loss for our framework.
	
	The normal accuracy can be improved by about 10\% when using the unsupervised pretext task in the STL-10 dataset.
	On the plankton dataset, the impact is less but tends to give better results of some percent.
	The pretext task in combination with the MI loss leads to a performance which is not more than 10\% worse than the full setup for all ablations except for one.
	For this exception, \ceinv is needed to stabilize the overclustering performance due to the poor initialization with CIFAR-20 weights.
	We attribute this worse performance to the initialization and not the different backbone because on STL-10 the CIFAR-20 initializations of the ResNet34 backbone outperform the ImageNet weights of the ResNet50v2 backbone.
	We believe the positive effects of ImageNet weights for its subset STL-10 and the better network are negated by the different loss.
	Overall, our novel loss \ceinv improves the overclustering performance regardless of the dataset and the weight initialization.
	The improvements are 10\% on STL-10, up to 7\% on the plankton dataset.
	
	On the normal head, IIC is one to two percent better or worse than the full FOC depending on the dataset.
	IIC is similar to FOC with pretext task and no additional losses.
	It outperforms this similar and restricted version of FOC by up to 10\% on the normal head.
	Since we train a more complex model with overclustering heads, we need the additional losses to reach a similar performance on the normal head.
	However, we additionally get the overclustering heads.
	We have seen above that overclustering is the key to handling fuzzy data and identifying substructures.
	Taking the overclustering results into consideration, we achieve an F1-Score of more than 11\% better in comparison to IIC.
	Even a boosted IIC with MLP heads instead of linear layers is around 8\% worse than our framework.
	
	\tblAblation

	\section{Conclusion}
	
	Current state-of-the-art semi-supervised learning algorithms allow training with 10 times fewer labels.
	In this paper, we take the first steps to address real-world issues with semi-supervised learning.
	Our presented novel framework FOC can handle fuzzy labels via overclustering.
	We illustrated the performance gain of our method in comparison to previous work in regard to runtime and overclustering accuracy on STL-10.
	Our framework was applied to a real-world plankton dataset with fuzzy labels.
	On fuzzy data, we showed that overclustering can achieve better results than common classification with normal heads.
	We reached an F1-score of about 78\% on the fuzzy data and illustrated the visual similarity on qualitative results from these predictions.
	This performance is greater than state-of-the-art semi-supervised frameworks like FixMatch and results in 5 to 10\% more consistent predictions.
	
	\section*{Acknowledgments}
	
	We thank our colleagues, especially Claudius Zelenka, for their helpful feedback and recommendations on improving the paper. 
	Moreover, we are grateful for all citizen scientist which participated in PlanktonID and the team of PlanktonID for providing us with their data.
	We thank Xu Ji, Ting Chen, Kihyuk Sohn and Wouter Van Gansbeke for answering our questions regarding their respective work.

	{\small
		\bibliographystyle{ieee_fullname}
		\bibliography{references}
	}

\end{document}